%% file: main.tex
\documentclass[10pt,twocolumn,letterpaper]{article}

\usepackage{iccv}              %

\definecolor{iccvblue}{rgb}{0.21,0.49,0.74}
\usepackage[pagebackref,breaklinks,colorlinks,allcolors=iccvblue]{hyperref}

\usepackage{bm}
\renewcommand{\paragraph}[1]{\noindent\textbf{#1}}

\def\confName{ICCV}
\def\confYear{2025}

\title{\ours: Unifying Knowledge %
from Large-scale and Diverse Pre-trained Models 
}

\author{Yimu Wang$^{\dag}$, Weiming Zhuang$^{\ddagger}$, Chen Chen$^{\ddagger}$, Jiabo Huang$^{\ddagger}$, Jingtao Li$^{\ddagger}$, Lingjuan Lyu$^{\ddagger}$\\
$^{\dag}$University of Waterloo, $^{\ddagger}$SONY AI
}

\input{utils}

\begin{document}
\maketitle

\begin{abstract}
In the era of deep learning, 
the increasing number of pre-trained models 
available online presents a wealth of knowledge. 
These models, developed with diverse architectures and 
trained on varied datasets for different tasks, 
provide unique interpretations of the real world. 
Their collective consensus is likely universal and generalizable to unseen data. 
However, effectively harnessing this collective knowledge 
poses a fundamental challenge 
due to the heterogeneity of pre-trained models. 
Existing knowledge integration solutions 
typically rely on strong assumptions about training data distributions and network architectures, 
limiting them to learning only from specific types of models and resulting in data and/or inductive biases.
In this work, we introduce a novel framework, namely \ours, for 
knowledge transfer from a diverse set of off-the-shelf models 
into one student model
without such constraints. 
Specifically, 
we propose a dedicated voting mechanism to capture the consensus of knowledge 
both at the logit level --
incorporating teacher models that are capable of predicting target classes of interest --
and at the feature level, utilizing visual representations learned on arbitrary label spaces.
Extensive experiments demonstrate that 
\ours effectively enhances unsupervised object recognition performance 
compared to strong knowledge transfer baselines. 
Notably, it exhibits remarkable scalability by benefiting from over one hundred teachers, 
while existing methods saturate at a much smaller scale.

\end{abstract}

\begin{figure}[t!]
\centering
\includegraphics[width=1\linewidth]{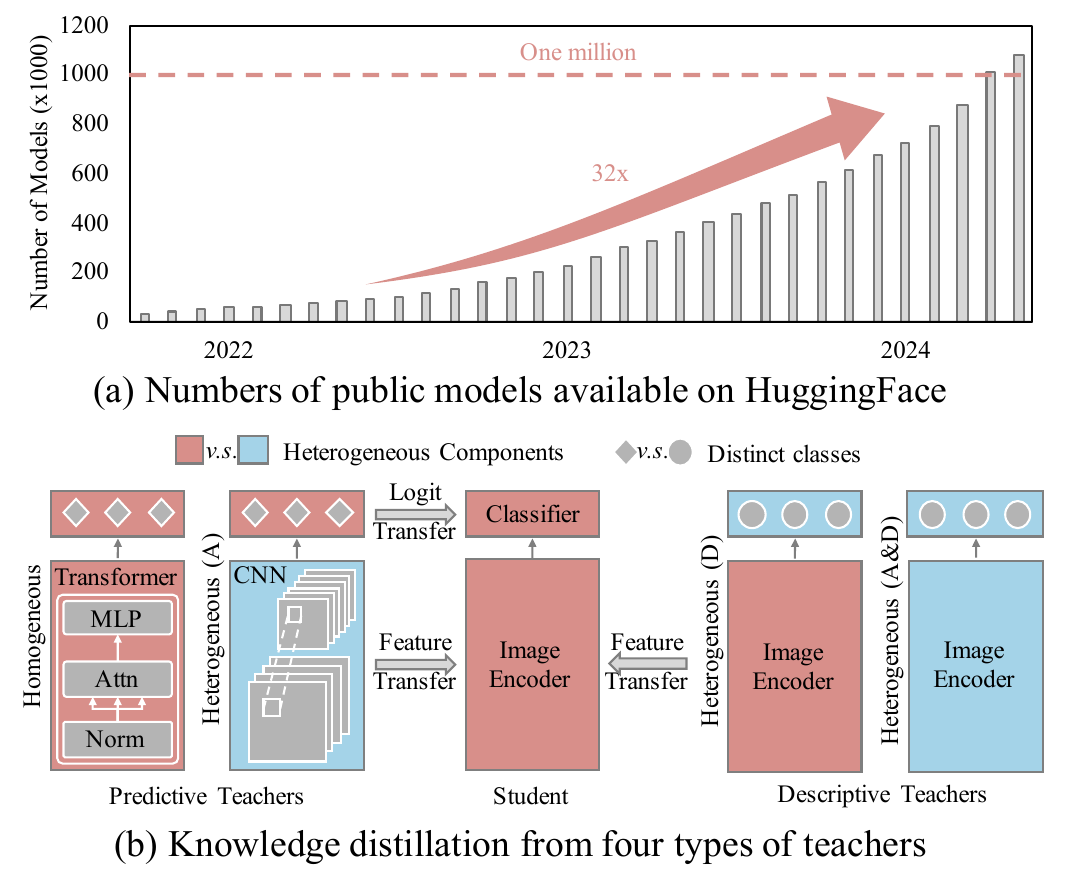}
\caption{
(a) The number of publicly available models
increase dramatically in the past few years
and exceeded one million now.
(b) %
These models serve as a valuable source of knowledge
for generic visual perception.
In this work, we introduce \ours to train a strong vision model through knowledge transfer from both predictive and descriptive teachers. These teachers include not only homogeneous models—those sharing the same architecture and label space as the student—but also heterogeneous models trained on distinct data classes (D), with different architectures (A), or varying in both aspects (A\&D).
}\label{fig: teaser}
\vspace{-1em}
\end{figure}

\section{Introduction}

Recent years have witnessed the rapid progress of deep learning~\cite{devlinBERTPretrainingDeep2019,dosovitskiyImageWorth16x162021,heDeepResidualLearning2016}. 
As depicted in \cref{fig: teaser} (a),
the number of pre-trained models available online has been dramatically increasing
since the introduction of HuggingFace~\cite{wolf-etal-2020-transformers} and similar platforms~\cite{noauthor_papers_nodate}. 
By building with different training techniques on varied data,
these off-the-shelf models offer diverse perspectives on understanding the world~\cite{li_implicit_2019,wangImplicitBiasAdaptive2021,wang_momentum_2022}. 
Whilst such rich visual knowledge from model zoos
can serve as an alternative and/or complement to manual labels
for training a strong and generalizable vision model,
this is challenging
due to the inconsistent representations of knowledge
produced by different models.

Model merging~\cite{akiba_evolutionary_2024,wang_sam-clip_2023,yadav_ties-merging_2023}
is a straightforward solution
for knowledge integration from multiple pre-trained models
by fusing their weights.
However, it is limited to benefiting from models with identical architectures
whose visual knowledge is 
constrained by the inductive bias of network designs.
Mixture-of-experts~\cite{lin_moe-llava_2024,satar_rome_2022,shen_scaling_2023,wu_omni-smola_2024} is another popular solution in recent years. 
It assembles models by routing among their submodules, whose high demands for storage and computational resources make it less applicable in some practical scenarios with strained budgets. 
On the other side, knowledge distillation (KD)~\cite{hinton_distilling_2015,zhang_task-oriented_2020,zhao_decoupled_2022} transfers ``dark knowledge'' from teachers to student by aligning 
either their prediction scores (\ie logits)~\cite{hinton_distilling_2015,zhang_task-oriented_2020,zhao_decoupled_2022,peng_correlation_2019,romero_fitnets_2015,tian_contrastive_2022} 
or latent embeddings (\ie features)~\cite{romero_fitnets_2015,du_agree_2020,heo_comprehensive_nodate,peng_correlation_2019,tian_contrastive_2022}.
However, the pre-trained models available in public model zoos are extremely diverse,
which could be built with arbitrary network designs on any data.
Existing KD methods either assume identical training distributions
to enable logit transfer~\cite{hinton_distilling_2015} which is subject to data bias,
or conducts feature transfer solely~\cite{ranzinger2024radio,sariyildiz2024unic} 
and overlooks the high-level semantics encoded in the teacher's class predictions.

In this work,
we address the challenge of 
effectively harnessing the rich visual knowledge
from the collective consensus of freely accessible model zoos,
and apply it to object recognition with unlabeled images. 
To this end, we propose \textbf{\ours}, 
a \textsc{unif}ied kn\textsc{o}wledge t\textsc{r}ansfer fra\textsc{m}ework capable of 
learning from a large collection of diverse teacher models with minimal constraints. 
Given the implementations of public models%
\footnote{Implementations may vary from simple model cards to fully detailed codebases.} 
and the target classes of interest, 
we categorize teachers into two groups, as shown in \cref{fig: teaser} (b).
We define \textit{predictive teachers} as 
models trained on a label space aligned with the target classes,
\eg source models in domain adaptation~\cite{wangDeepVisualDomain2018}, 
open-vocabulary classifiers~\cite{radford_learning_2021,girdhar_imagebind_2023}, 
or even “black-box” APIs providing target predictions without revealing underlying methods. 
In contrast, 
we refer to models that do not share the target classes but 
provide informative visual feature representations as \textit{descriptive teachers}.
For effective knowledge integration and transfer from both types of teachers, we introduce novel voting mechanisms that allow teachers to regularize each other by exploring their consensual knowledge. 
Specifically,
we map all teachers' features to a unified space
and filter potentially biased information by solving the sign conflicts in features as shown in \cref{fig: FT}.
This design enables voting of features that are pre-learned in disjoint latent spaces by different teachers,
and alleviates noisy supervision in student's feature learning.
Additionally, 
we adopt class prediction voting across all predictive teachers
to estimate a reliable pseudo-label
to be emphasized in logit transfer,
so as to mitigate the potential confusion caused by
the inconsistent predictions
from different teachers as shown in \cref{fig: LT}.

In summary, our contributions are in three-folds:
\textbf{(1)} We explore the scalability of knowledge transfer from over one hundred publicly available models 
built with diverse network architectures and trained on various data domains. 
Our results underscore the substantial potential of 
freely accessible knowledge within public model zoos.
\textbf{(2)} We propose \ours, a unified knowledge transfer framework, 
to integrate knowledge from a set of off-the-shelf models 
as an alternative supervision to manual labels 
for training a strong object recognition model.
\ours is characterized by dedicated voting mechanisms which
refrain the student model from the potentially biased knowledge 
and ensure its generalizability by only learning from 
the collective consensus of teachers.
\textbf{(3)} Extensive experiments show the superior performance of \ours over 
several strong KD baselines
on 11 object recognition benchmark datasets,
and sometimes even outperform the teachers.
Moreover, \ours demonstrates superior scalability to the number of teachers
while existing KD methods saturate at a much smaller scale.

\section{Related Work}

\noindent
\textbf{Model merging.} 
Merging different models becomes a practical solution for improving the performance on a single target task~\cite{choshen_fusing_2022,gupta_stochastic_2020,izmailov_averaging_nodate,wortsman_model_2022}, enhancing out-of-domain generalization~\cite{arpit_ensemble_2022,cha_swad_2021,ilharco_editing_2022,jin_dataless_2022,rame_diverse_2022,rame_model_2023}, creating multitask models from different tasks~\cite{li_branch-train-merge_2022}, compression~\cite{li_merge_2023}, multimodal understandings~\cite{sung_empirical_2023}, continual learning~\cite{yadav_exclusive_2023}, \etc
Whilst being shown effective,
it requires candidate
models (models to be merged) to share the same architectures.
However, public models online are usually trained with varied designs.
\ours transfers knowledge from the model's inference behaviors rather than adapting their pre-trained weights,
which poses no constraints on network architectures.

\paragraph{Mixture of experts (MoE)}
is a hybrid model consisting of multiple experts~\cite{eigen_learning_2013} sub-models.
Each expert
has its unique set of weights, enabling them to craft distinct representations for each input token based on contextual information. 
The key concept is the use of a router to determine the token set that each expert handles, thereby reducing interference between different types of samples~\cite{lin_moe-llava_2024,satar_rome_2022,shen_scaling_2023,wu_omni-smola_2024}.
MoE needs all the models to be stored and run at inference. 
This is less practical in some real-world scenarios with limited storage or computational budgets like pedestrian surveillance in edge systems~\cite{li2018harmonious}.
Our \ours integrates the knowledge of different teachers into a single student model which can be built with any architecture that fits the target use cases.

\paragraph{Knowledge distillation} (KD)~\cite{hinton_distilling_2015}, as a representative approach for inheriting knowledge from off-the-shelf models by training a student model to mimic their inference behaviors~\cite{devlinBERTPretrainingDeep2019,dosovitskiyImageWorth16x162021,heDeepResidualLearning2016}, has attracted extensive attention in the last few decades. 
Logit-based KD methods 
take the teacher's predictions or responses as the soft labels for training the student.
For example, DML~\cite{zhang_deep_2018} trains students and teachers simultaneously via mutual learning.
TAKD~\cite{mirzadeh_improved_2020} introduces an intermediate-sized network to serve as a bridge
between teachers and students. 
Whilst logit-based methods have been shown effective in transferring class-specific semantic knowledge,
they are limited to benefiting from teachers sharing an identical label space with the student
and fails to exploit the generic visual knowledge derived from diverse data.
On the other side, feature-based methods directly transfer feature representations from teachers to students.
As a pioneer, FitNets~\cite{romero_fitnets_2015} is proposed to employ a convolutional layer to project student features to match the teacher's feature size and then align the features in the teacher's latent space. 
A similar idea has been further explored later on with more delicate designs~\cite{chen_distilling_2021,hao_learning_2022,zhang_task-oriented_2020}.
While feature-based methods achieve remarkable performance compared with logit-based methods, 
they fail to explore the high-level semantics embedded in the class predictions.
\ours conducts both logit and feature transfers 
and allows teachers to contribute partially to these objectives.
Such a flexible design enables us to take advantage of
visual knowledge at different levels from diverse teachers.

\section{Methodology}

\paragraph{Problem Definition.}
Given a set of $N$ \textit{unlabeled} images $\{I_i\}_{i\in[N]}$
drawn from a set of $C$ object classes,
we aim at learning a feature encoder $f_\theta(I_i) \rightarrow \bm{x}_i \in \mathbb{R}^{D}$,
where $D$ is the feature dimension,
along with a classifier $f_\phi(\bm{x}_i)\rightarrow \bm{p}_i$ so that
\begin{equation}
    \small
y_i = \arg\max_{c\in [C]} \bm{p}_i
\label{eq: predict}
\end{equation}
reveals the underlying semantic class of the $i$-th image.
The symbols $\theta$ and $\phi$ denote the learnable parameters of the target model.
Considering that the manual class labels are inaccessible
while there are numerous pre-trained models available online
providing rich visual knowledge,
we propose to train our model
by knowledge integration and transfer from these off-the-shelf models.
To this end,
we collect $N^t=N^p+N^d$ pre-trained models from publicly accessible model zoos~\cite{noauthor_papers_nodate,wolf-etal-2020-transformers},
consisting of
$N^p$ \textit{predictive teachers} 
producing both the feature representations $\{\bm{x}^{tp}_{i,j}\}_{j=1}^{N^p}$ of sample $I_i$
and their corresponding class predictions $\{\bm{p}^{tp}_{i,j}\}_{j=1}^{N^p} \in [0, 1]^{N^p\times C}$,
as well as $N^d$ \textit{descriptive teachers}
yielding image features $\{\bm{x}^{td}_{i,j}\}_{j=1}^{N^d}$ solely.
For clarify,
we omit the sample index and aggregate the features obtained from both types of teachers as
$\{\bm{x}^t_{i}\}_{i=1}^{N^t}$ and the class predictions from predictive teachers as
$\{\bm{p}^t_{i}\}_{i=1}^{N^p}$.
Likewise,
the student's features and logits are simplified to be
$\bm{x}$ and $\bm{p}$, respectively.

\subsection{\ours Knowledge Transfer}\label{sec: our method}

\begin{figure}
    \centering
    \includegraphics[width=\columnwidth]{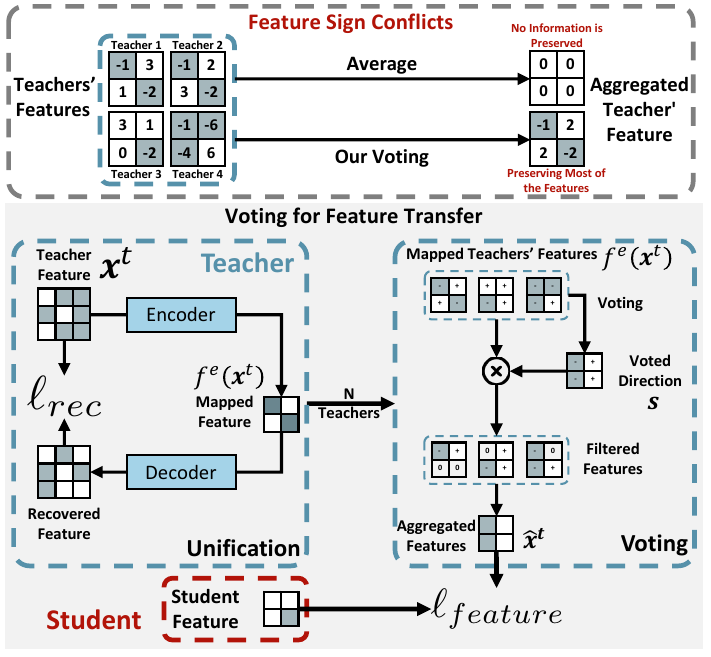}
    \vspace{-2em}
    \caption{
    \textbf{Sign conflicts (top).} 
    A simple average of teachers' features tend to result in less informative supervision due to their sign conflicts,
    \eg the agreed features can be all $0$ in an extreme case when all the teachers' features offset with each other.
    \textbf{Features voting and transfer (bottom).}
    We propose a two-stage framework for mitigating sign conflicts, which first unifies the features into a common space shared with the student, votes and conducts aggregation among mapped teachers' features, and then minimizes the distance between the student feature and the aggregated features. 
    During the voting procedure, the sign conflicts between teachers' features are resolved by element-wisely filtering out the features that do not have the same direction as most of the features.
    }
    \label{fig: FT}
    \vspace{-1em}
\end{figure}

\begin{figure}
    \centering
    \includegraphics[width=\columnwidth]{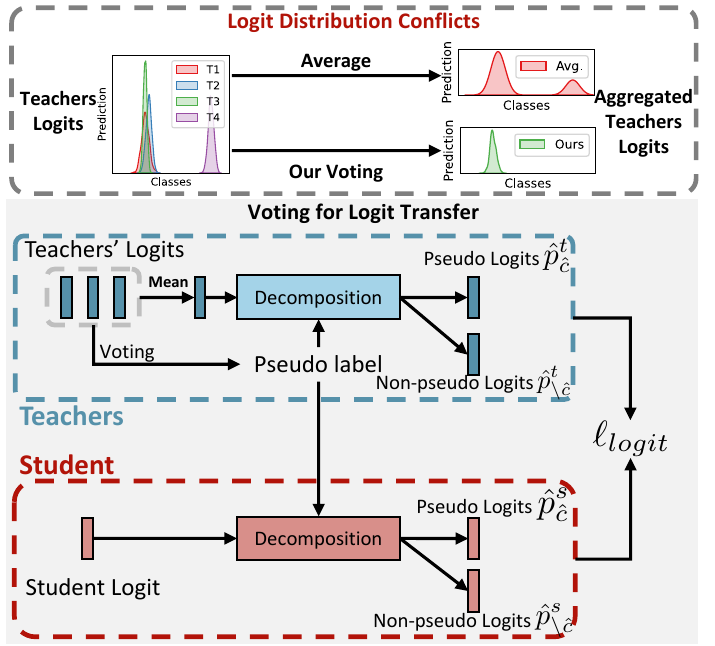}
    \vspace{-2em}
    \caption{
    \textbf{Logit distribution conflicts (top).} 
    As teachers usually produce inconsistent prediction distributions, it might confuse the student when directly forcing the student to mimic the behavior of all the teachers. 
    \textbf{Logits voting and transfer (bottom).}
    Instead of averaging on the teachers' logits, \ours highlights transfer on the pseudo-class voted by teachers to avoid confusion.
    \ours first gets the pseudo labels for the unlabeled public data by voting among teachers, decomposes the student and teachers' logits into pseudo and non-pseudo logits, and then processes knowledge transfer separately to highlight the importance of pseudo labels.
    }
    \vspace{-1em}
    \label{fig: LT}
\end{figure}

To summarize the treasured ``dark knowledge'' from public online teachers, we propose a unified cross-architecture and cross-training data knowledge integration method, namely \ours. 
It harnesses the rich knowledge from teacher models of any architecture trained on any benchmark dataset. 
However, due to the bias introduced by different training data and the heterogeneity of architecture, directly learning from diverse and even controversial/noisy teachers' knowledge becomes a challenge for the student model. 
To this end, we design two special voting mechanisms to encourage the student to imitate the teachers' features (\Cref{sec: feature learning} and \cref{fig: FT}) and logits (\Cref{sec: logit learning} and \cref{fig: LT}).

\subsubsection{Features Voting and Transfer}\label{sec: feature learning}

Learning from the visual features produced by teacher models has been proven effective~\cite{du_agree_2020,heo_comprehensive_nodate,peng_correlation_2019,romero_fitnets_2015,tian_contrastive_2022}
in the knowledge transfer literature.
However,
due to the diverse training data domains used to build different teachers and 
the potential distributional gaps between these domains and the target, 
teachers may provide noisy supervision.
It further yields lower-quality feature representations for 
samples that are out-of-distribution relative to their training data.
To mitigate such misleading signals, 
we propose a voting mechanism that filters potentially unreliable information in 
feature representations by identifying sign conflicts between different teachers' features.

\paragraph{Features unification.}
To effectively explore the consensus among teachers at the feature level, 
we first unify their features,
which are drawn from disjoint latent spaces learned independently,
into a common space shared with the student. 
By projecting teachers' features into a common latent space,
we are able to explicitly explore the potential correlations between different teachers,
whereas existing feature-based KD methods~\cite{ranzinger2024radio, sariyildiz2024unic} 
typically align the student's features with each teacher independently.
Specifically, for each teacher, we use a encoder $f^{e}_i(\cdot)$, where ${i \in [N^t]}$, to map its features 
into a $D$-dimensional latent space
to interact with other teachers and align with the student.
The mapped teachers' features are denoted as $\{f^e_i (\bm{x}^t_i)\}_{i\in[N^t]}$. 
To refrain from feature degeneration,
we employ decoders $\{f^{d}_i(\cdot)\}_{i \in [N^{t}]}$ to regularize the mapped features by a reconstruction loss $\ell_{rec}$ as follows,
\begin{equation}
    \small
    \ell_{rec} = \sum_{i \in [N^t]} \lVert f^{d}_i(f^{e}_i(\bm{x}^t_i)) - \bm{x}^t_i \rVert_{2}\,.
\end{equation}

\paragraph{Features voting.}
The simplest way to align the student features with the teachers' features is to take the average of the teachers' features and 
minimize its distance to the student's features.
However, as shown in \Cref{fig: FT}, due to the inductive bias brought by the heterogeneity of architecture and training data, the distributions of teachers' features differ as their signs might conflict with each other. 
That means by simply taking the average on the teachers' features, we might obtain an average feature whose most of the elements would be 0, which does not provide any information. 
To efficiently address this issue, 
we propose to mitigate the sign conflicts by voting among all the teachers to find the voted directions: 
\begin{equation}
    \small
    \bm{s} = \operatorname{sgn}\left(\sum_{i \in [N_{teachers}]} \operatorname{sgn}(f^{e}_i(\bm{x}_i^t))\right)\,,
\end{equation}
where $\operatorname{sgn}(\cdot)$ is the element-wise sign function
while $\bm{s}$ is a $D$-dim vector indicating the agreement at each feature dimension.

We then filter out the elements that do not agree with the (voted) consensus directions 
and aggregate the remaining features by,
\begin{equation*}
    \scriptsize
    \hat{\bm{x}}^{t} =  \frac{\sum_{i \in [N^t]} \mathbb{1}(\operatorname{sgn}(f^{e}_i(\bm{x}_i^t)) = \bm{s})\odot f^e_{i}(\bm{x}_i^t) }{\sum_{i \in [N^t]} \mathbb{1}(\operatorname{sgn}(f^e_{i}(\bm{x}_i^t)) = \bm{s})}\,,
\end{equation*}
where $\mathbb{1}(\cdot)$ is the indicator function and $\odot$ stands for the hadamard product.
After obtaining the filtered features, to let the student mimic teachers' behavior, we minimize the following losses, 
\begin{equation}
\label{eqn: feature loss}
    \small
    \ell_{feature} = \operatorname{dist} (\bm{x}, \hat{\bm{x}}^t)\,,
\end{equation}
where $\operatorname{dist} (\cdot, \cdot)$ is a distance metric, \eg, maximum mean discrepancy~\cite{grettonKernelTwoSampleTest2012}, $\ell_2$ distance, or Kullback–Leibler divergence.

\subsubsection{Logits Voting and Transfer}\label{sec: logit learning}

Due to the inductive biases of models brought by network architectures,
different models trained on the same label space or even using the same training data
can lead to inconsistent prediction distributions~\cite{allen-zhu_towards_2022,biettiInductiveBiasNeural2019}.
For example, CNN models~\cite{heDeepResidualLearning2016,simonyanVeryDeepConvolutional2014,zagoruykoWideResidualNetworks2016} focus more on local details, while vision transformer models~\cite{dosovitskiyImageWorth16x162021,wang_seggpt_2023} tend more to global information with the introduction of the attention mechanism~\cite{vaswaniAttentionAllYou2017}. 
This difference results in logits inconsistency as shown in \cref{fig: LT}, 
which might further confuse the students during training if we let the average of teachers' predictions as the target. 

\paragraph{Logits voting.}
To alleviate the confusion brought by inconsistent class predictions from different teachers, 
instead of treating all classes equally in logit transfer,
we want to highlight the ones that are more likely to be true
and alleviate the impacts of the rest which tend to be distracting.
Therefore,
we estimate the most likely class label
by teacher voting following
\begin{equation}
    \small
\begin{gathered}
\hat{c} = \arg\max \tilde{p}, \text{where} \\
\tilde{p} = \{\frac{\sum_{i\in[N^p]}\mathbb{1}[\arg\max_{j\in[1,C]}p^t_{i,j}=c]}{N^p} \vert \forall c\in[C]\}.
\end{gathered}
\label{eq: pseudo class}
\end{equation}
In \cref{eq: pseudo class},
$p^t_{i,j}$ denotes the predicted score made by the $i$-th teacher on the $j$-th class,
and $\hat{c}$ is considered as the pseudo-class that the input sample is most likely coming from. 
We then aggregate different teachers' logits by average pooling,
$\hat{\bm{p}}^t = \frac{\sum_{i \in [N^p]}\bm{p}^t_i}{N^p}$. 
Here, we assume predictive teachers share the same label space for simplicity. 
For implementation, the predictions are averaged class-wisely among the teachers that can predict those classes.
Inspired by \citet{zhao_decoupled_2022},
we emphasize the pseudo-class when transferring knowledge at the logits level:
\begin{equation}
\small
\begin{aligned}
    \ell_{logit} = 
    & \underbrace{H(\hat{p}^{t})}_{\text{constant}} + 
    \underbrace{\alpha_1(\hat{p}_{\hat{c}}^{t} \log {p}_{\hat{c}})}_{\text{pseudo class}} +   \underbrace{\alpha_2(\sum_{c\in[1,C],c\neq\hat{c}}\hat{p}_{c}^{t} \log {p}_{c})}_{\text{non-pseudo classes}} \,,
\end{aligned}
\end{equation}
where $\alpha_1$ and $\alpha_2$ are the hyperparameters
for balancing the importance of pseudo and non-pseudo classes.

\subsubsection{Model Training and Inference}

With our joint feature and logit transfer designs
as well as the dedicated voting mechanisms,
\ours is capable of leveraging extensive pre-trained models as its teachers
and learn to recognize visual objects in images
without manual labels. 
The overall learning objective of \ours is:
\begin{equation}
    \mathcal{L} = \ell_{logit} + \beta_1 \ell_{feature} + \beta_2 \ell_{rec}\,,
\end{equation}
where $\beta_1$ and $\beta_2$ are hyperparameters.
After training, 
all the teacher models are deprecated
and only the student model will be used to make predictions according to \cref{eq: predict}.

\section{Experiments}

\paragraph{Benchmark datasets.} 
We employ a total of 11 datasets for evaluating the performance of \ours, %
\ie, CUB200~\cite{wah_notitle_2011}, Stanford Dog~\cite{khoslaNovelDatasetFinegrained2011}, Flowers102~\cite{nilsback_automated_2008}, OxfordPet~\cite{parkhi_cats_2012}, Stanford Cars~\cite{krause3DObjectRepresentations2013}, Caltech101~\cite{li_caltech_2022}, Cifar10~\cite{krizhevskyLearningMultipleLayers2009}, Cifar100~\cite{krizhevskyLearningMultipleLayers2009}, DTD~\cite{cimpoi_describing_2014}, Aircraft~\cite{majiFineGrainedVisualClassification2013}, and Food101~\cite{bossardFood101MiningDiscriminative2014}. 

\noindent
\textbf{Evaluation Protocol. }
Considering that it is sometimes challenging
to collect sufficient predictive teachers 
which are trained on the exact target label space,
we evaluate \ours in a more practical scenario
where each predictive teacher
can only make predictions on a subset of the target classes and $\hat{\bm{p}}^t$ is calculated by class-wisely averaging the teachers' predictions.
To address this,
we combine multiple datasets
and conduct knowledge transfer to
train our student model on the union of their label spaces.
Specifically, if the label space of the $i$-th dataset is $\mathcal{Y}_i = \{y_{c}\}_{c\in[C_i]}$ and we have $K$ datasets in total, 
the union of label spaces is $\mathcal{Y}_1 \cup \cdots \cup \mathcal{Y}_{K}$ and 
we train a model
to classify images into one of
$C = \sum_{i\in[K]} C_i$ classes. 
All the samples in the datasets selected for combination will be used for training.

\noindent
\textbf{Evaluation metrics.} 
We report the accuracy of each dataset unless otherwise specified. 
Moreover, we do not have or assume any prior during the training and inference (classification).
During training, our voting mechanism selects the most likely class, while during inference, the classification is conducted on the union of label spaces. 
For example, if a ``Bobolink'' image from CUB200 has the biggest probability of ``Husky'' (a class in Dog) and the second highest probability of ``Bobolink'' (a class in CUB200), we will mark it as misclassified into ``Husky''. 
In this case, we report the average accuracy by datasets and classes by Avg. (D) and Avg. (C), respectively. 
The average accuracy by datasets is calculated as the mean of the accuracies across all datasets. In contrast, the average accuracy by classes is determined by averaging the accuracy of each class across all datasets.

\begin{table*}[t!]
\centering
\begin{tabular}{l|ccc|cc|cc}
\toprule
\multirow{2}{*}{Methods}                          & \multirow{2}{*}{\shortstack{Labeled\\Data?}}          & \multirow{2}{*}{\shortstack{Predictive\\ Teachers}} & \multirow{2}{*}{\shortstack{Descriptive\\ Teachers}} & \multicolumn{4}{c}{Accuracy (\%)}                                                                          \\
&                         &             &               & CUB200                                 & Dogs                             & \textbf{Avg. (D)} & \textbf{Avg. (C)} \\ \midrule

\color{gray}{Predictive Teacher (ViT)}           &     $\checkmark$ &                                                                                                                                 &                                                                & \color{gray}{86.59} & \color{gray}{85.90} & -        & -        \\ 
\midrule
KD~\cite{hinton_distilling_2015}   &$\times$ & $\checkmark$                                                                                                                     &                                                                &     85.31 & 87.21 & 86.21 & 85.90       \\
CFL~\cite{luo_knowledge_2019}   &$\times$    & $\checkmark$                                                                                                             &                                                                &    85.62 & 88.64 & 87.12 & 86.64    \\
OFA~\cite{hao_one-for-all_2023} &$\times$ & $\checkmark$                                                                                                          &                                                                &   83.09 & 90.44 & 86.76 & 85.61       \\
\midrule
CFL+&$\times$ & $\checkmark$                                                       & $\checkmark$                                                         &     85.93 & 88.86 & 87.39 & 86.85  \\
\rowcolor{green!10} \ours &$\times$& $\checkmark$                                                       & $\checkmark$    &\textbf{86.52} & \textbf{90.65} & \textbf{88.58} & \textbf{87.94} \\
\bottomrule
\end{tabular}%
\vspace{-1em}
\caption{
Experimental results on transferring knowledge from both predictive and descriptive teachers to the student (ViT) on \textbf{2 datasets}, namely CUB200 and Stanford Dogs. 
Avg. (D) and Avg. (C) refer to the average accuracy of datasets and classes. 
Predictive teachers are individually trained and tested on each dataset. 
Thus, as \ours and other KD methods are trained on the joint label spaces and training data, we do not present the Avg. (D) and Avg. (C) results of predictive teacher for a fair comparison. 
Best in \textbf{Bold}.
}
\label{tab: ablation: 2 datasets}

\centering
\resizebox{\textwidth}{!}{%
\begin{tabular}{l|ccc|ccccc|cc}
\toprule
\multirow{2}{*}{Methods} & \multirow{2}{*}{\shortstack{Labeled\\Data?}} & \multirow{2}{*}{\shortstack{Predictive\\ Teachers}} & \multirow{2}{*}{\shortstack{Descriptive\\ Teachers}} & \multicolumn{7}{c}{Accuracy (\%)}                                             \\
                         &                         &             &                                 & CUB200 & Flowers102 & Pets & Cars & Dogs & \textbf{Avg. (D)} & \textbf{Avg. (C)} \\
\midrule
\color{gray}{Predictive Teacher (ViT)}                       &    $\checkmark$&                                                                     &                                  &    \color{gray}{86.59} &    \color{gray}{96.86} &  \color{gray}{93.13} &    \color{gray}{82.80} &    \color{gray}{85.90}   &-&-  \\ 
\midrule
KD~\cite{hinton_distilling_2015}                       &  $\times$&   $\checkmark$                                                                    &                                  &   85.26 & 95.85 & 89.42 & 71.76 & 67.83 & 82.03 & 80.05     \\ 
CFL~\cite{luo_knowledge_2019}                         &  $\times$&        $\checkmark$                            &        & 84.81   &      95.61      &     90.90        &       72.71        &      71.83         &  83.17 &  80.92  \\
OFA~\cite{hao_one-for-all_2023}                         &  $\times$&       $\checkmark$                              &                                  &   86.21 & 97.89 & 90.98 & 74.70 & 73.36 & 84.62 & 82.57  \\ 
\midrule
CFL+                         &  $\times$&         $\checkmark$                           &      $\checkmark$                     &  85.69      &      95.67      &     93.51        &        72.70       &         \textbf{88.46}      & 87.21  & 84.39   \\
\rowcolor{green!10} \ours                         &       $\times$&    $\checkmark$                           &           $\checkmark$           &  \textbf{86.43} & \textbf{98.11} & \textbf{93.68} & \textbf{77.10} & 88.40 & \textbf{88.75} & \textbf{86.15} \\
\bottomrule
\end{tabular}
}
\vspace{-1em}
\caption{
Experimental results on transferring knowledge from predictive and descriptive teachers to the student (ViT) on \textbf{5 datasets}. 
``Pets'', ``Cars'', and ``Dogs'' are OxfordPet, Stanford Cars, and Dogs. 
Avg. (D) and Avg. (C) refer to the average accuracy of datasets and classes. 
}
\label{tab: main 5 datasets}
\centering
\resizebox{\textwidth}{!}{%
\begin{tabular}{l|ccccccccccc|cc}
\toprule
\multirow{2}{*}{Methods}                           & \multicolumn{13}{c}{Accuracy (\%)}                                                                                                                                                                                                                   \\
                                                   & CUB                                 & Flow.                             & Pets                                   & Cars                                   & Dogs                                   & Food & Calt. & C10 &C100 & DTD & Air. & \textbf{Avg. (D)} & \textbf{Avg. (C)} \\
                                                   \midrule
\color{gray}{Predictive Teacher (ViT)}  & \color{gray}{86.59} & \color{gray}{96.86} & \color{gray}{93.13} & \color{gray}{82.80} & \color{gray}{85.90} & \color{gray}{87.47} & \color{gray}{94.72} & \color{gray}{98.48} & \color{gray}{91.14} & \color{gray}{74.10} & \color{gray}{61.09}  & -        & -       
\\
\midrule
KD~\cite{hinton_distilling_2015}  & 78.62 & 94.44 & 91.69 & 53.64 & 84.16 & 81.17 & 89.15 & 96.98 & 85.90 & 67.07 & 45.24 & 78.91 & 75.19 \\
CFL~\cite{luo_knowledge_2019}   & 79.63 & 94.45 & 91.47 & 56.42 & 83.83 & 80.37 & 90.92 & 96.34 & 85.12 & 67.23 & 43.44 & 79.02 & 75.58 \\
OFA~\cite{hao_one-for-all_2023} & 82.36 & 96.94 & \textbf{92.97} & 55.53 & 89.20 & \textbf{85.40} & 90.67 & 97.72 & \textbf{89.06} & 69.89 & 45.63 & 81.39 & 78.03 \\
\midrule
CFL+ & 81.84 & \textbf{96.96} & 92.80 & 54.17 & \textbf{89.24} & 85.28 & 90.52 & \textbf{97.80} & 89.00 & \textbf{70.37} & 44.58 & 81.14 & 77.61 \\
\rowcolor{green!10} \ours & \textbf{82.95} & 96.83 & 92.23 & \textbf{66.61} & 86.93 & 84.84 & \textbf{91.50} & 95.79 & 88.54 & 69.96 & \textbf{53.62} & \textbf{82.87} & \textbf{80.47} \\
\bottomrule
\end{tabular}%
}
\vspace{-1em}
\caption{
Experimental results on transferring knowledge from both predictive and descriptive teachers to the student (ViT) on \textbf{11 datasets}. 
``Flow.'', ``Food'', ``Calt.'', ``C10'', ``C100'', and ``Air.'' represents Flowers102, Food101, Caltech101, Cifar10, Cifar100, and Aircraft. 
}
\label{tab: ablation: 10 datasets}
\end{table*}

\noindent
\textbf{Baselines.} 
As our method is the first to employ teachers with diverse architecture and training data at both feature and logit levels, 
for comparison with KD methods, we have set up different baselines as follows,
(1) methods with predictive teachers only: Knowledge Distillation (KD)~\cite{hinton_distilling_2015}, CFL~\cite{luo_knowledge_2019} and OFA~\cite{hao_one-for-all_2023} without human labels, and 
(2) methods with predictive and descriptive teachers: since there does not exist a method of learning jointly from predictive and descriptive teachers, we extend CFL to support descriptive teachers, namely CFL+. 
The details of CFL+ are in the Appendix.

\noindent
\textbf{Public teachers.} 
We select teachers from HuggingFace trained on 11 benchmark datasets with the architecture being ResNet50~\cite{heDeepResidualLearning2016}, ViT-b32~\cite{dosovitskiyImageWorth16x162021}, ConvNeXt-base~\cite{liu_convnet_2022}, and SwinTransformer~\cite{liu_swin_2021}. 
Besides, 
we select 10 well-known and powerful pre-trained models from HuggingFace and PapersWithCode 
as well as 50 top downloaded models on HuggingFace
as the \textit{out-of-distribution} descriptive teachers. 
The details of public teachers can be found in the Appendix.

\noindent
\textbf{Implementation details.} 
We evaluated a variety of student models with different architectures. 
Specifically, we have ResNet50~\cite{heDeepResidualLearning2016}, ViT-b32~\cite{dosovitskiyImageWorth16x162021}, ConvNeXt-base~\cite{liu_convnet_2022}, and SwinTransformer~\cite{liu_swin_2021}. 
All the experiments run for 100 epochs on 2 H100 GPUs. 
The encoder $f_{*}^{e}$ and decoder $f_{*}^{d}$ consists of four 2D convolutional layers and ReLU activation. 
To improve the training efficiency, instead of generating teachers' features in an online manner at training, we pre-calculate teachers' features and load them during training. 
During training, only the teachers' encoders and decoders and the student model are optimized and loaded on GPUs.

\subsection{Main Results}

We conduct a comprehensive study of \ours's capability to
transfer knowledge from off-the-shelf models 
in different practical scenarios
by varying the number of datasets to be combined.
Specifically,
we first verify our model designs in a simple case
by combing two datasets which are distinct both visually and semantically.
We then investigate
how robust \ours is on the combination of 5 datasets
where different classes might be similar to a certain extent
and their corresponding predictive teachers are likely distracting to each other.
Lastly,
we evaluate \ours in the most challenging scenario
by combining all 11 datasets
with the largest and most complex target label space
while each predictive teacher covers only a limited subset.

\noindent
\textbf{Effectiveness of \ours}. 
We start with evaluating \ours on 
the combination of CUB200 and Stanford Dogs.
Due to their visually dissimilar images and semantically irrelevant classes,
predictive teachers are less likely to be distracted by each other
and result in unreliable supervision of the student.
As shown in \Cref{tab: ablation: 2 datasets},
CFL~\cite{luo_knowledge_2019} achieves the best performances
among the three logit-based KD approaches.
By extending CFL to incorporate feature transfer,
its performances are improved consistently on both the two datasets,
which demonstrates the benefits brought by descriptive teachers that trained on other label spaces.
Furthermore,
the notable performance advantages achieved by \ours over CFL+
verify the effectiveness of \ours on knowledge transfer from diverse public teachers.

\begin{table*}
\centering
\resizebox{\textwidth}{!}{%
\begin{tabular}{l|cc|ccccc|cc}
\toprule
\multirow{2}{*}{Methods} & \multirow{2}{*}{\shortstack{Logit Transfer\\Voting}} & \multirow{2}{*}{\shortstack{Feature Transfer\\Voting}} & \multicolumn{7}{c}{Accuracy}                                             \\
                         &                                                                             &                                  & CUB200 & Flowers102 & Pets & Cars & Dogs & Avg. (D) & Avg. (C) \\
                         \midrule
CFL+                         &                                   &                                  &  85.69      &      95.67      &     93.51        &        72.70       &         88.46      & 87.21  & 84.39   \\
\midrule
\multirow{3}{*}{\ours}&               $\checkmark$                                  &          &  86.47 & 97.98 & 93.57 & 76.55 & \textbf{88.40} & 88.59 & 85.97       \\
                         &                                                 &       $\checkmark$                     &86.16 & 97.87 &  93.21 & 76.12 & 88.33 & 88.33 & \textbf{86.71}\\
                         &               $\checkmark$                                  &       $\checkmark$  & \textbf{86.43} & \textbf{98.11} & \textbf{93.68} & \textbf{77.10} & \textbf{88.40} & \textbf{88.75} & 86.15
                         \\
\bottomrule
\end{tabular}%
}
\vspace{-1em}
\caption{
Ablation studies on different components. 
CFL+ serves as our base model.
}
\label{tab: ablation: components}
\centering
\begin{tabular}{lcc|ccccc|cc}
\toprule
\multirow{2}{*}{Methods} & \multirow{2}{*}{\shortstack{Predictive\\Teachers}} & \multirow{2}{*}{\shortstack{Descriptive\\Teachers}} & \multicolumn{7}{c}{Accuracy} \\
 & & & CUB200 & Flowers102 & Pets & Cars & Dogs & Avg. (D) & Avg. (C) \\
\midrule
CFL~\cite{luo_knowledge_2019}                        &        $\checkmark$                            &        & 84.81   &      95.61      &     90.90        &       72.71        &      71.83         &  83.17 &  80.92  \\
CFL+                         &         $\checkmark$                           &      $\checkmark$                     &  85.69      &      95.67      &     93.51        &        72.70       &         {88.46}      & 87.21  & 84.39   \\
\midrule
\multirow{2}{*}{\ours} & $\checkmark$ &  & 85.50 & 97.90 & \textbf{94.14} & 71.17 & \textbf{90.66} & 87.88 & 84.47 \\
&  $\checkmark$ & $\checkmark$ & {\textbf{86.43}} & {\textbf{98.11}} &{93.68} & \textbf{77.10} & {88.40} & \textbf{88.75} & \textbf{86.15} \\
\bottomrule
\end{tabular}%
\vspace{-1em}
\caption{
Ablation studies on different types of teachers. 
Avg. (D) and Avg. (C) refer to the average accuracy of datasets and classes. 
}\label{tab: ablation: different teachers}
\centering
\resizebox{\textwidth}{!}{%
\begin{tabular}{l|ccc|ccccccc}
\toprule
\multirow{2}{*}{Methods}      & \multirow{2}{*}{\shortstack{Labeled \\ Data?}}    & \multirow{2}{*}{\shortstack{Preditive and\\Descriptive Teachers}} & \multirow{2}{*}{\shortstack{Student\\ Architecture}} &  \multicolumn{7}{c}{Accuracy}\\
&  & &  & CUB200   & Flowers102  & Pets & Cars  & Dogs & Avg. (D) & Avg. (C) \\ \midrule
\color{gray}{Predictive Teacher }                                                                        &  $\checkmark$&-        & ViT                                                                     & \color{gray}{86.59} & \color{gray}{96.86} & \color{gray}{93.13} & \color{gray}{82.80} & \color{gray}{85.90} & -        & -        \\
CFL+ & $\times$& $\checkmark$                                                                                                                & ViT                                                                      & {85.69}      &      {95.67}      &     {93.51}        &        {72.70}       &         {\textbf{88.46}}      & {87.21}  & {84.39} \\
\rowcolor{green!10} \ours                                                        & $\times$&$\checkmark$                                                         & ViT                                                                      & {\textbf{86.43}} & {\textbf{98.11}} & {\textbf{93.68}} & {\textbf{77.10}} & 88.40 & {\textbf{88.75}} & {\textbf{86.15}}\\
\midrule
\color{gray}{Predictive Teacher }                                                                        &  $\checkmark$&-        & ResNet-50                                                                     &  \color{gray}{78.75} & \color{gray}{92.68} & \color{gray}{90.35} & \color{gray}{87.23} & \color{gray}{81.54}& -        & -   \\
CFL+&$\times$& $\checkmark$                                                                                                                & ResNet-50 & 78.32 & 89.43 & 90.54 & 81.17 & 70.40 & 81.97 & 80.52\\
\rowcolor{green!10} \ours   & $\times$&$\checkmark$                                                         & ResNet-50 & \textbf{78.68} & \textbf{89.64} & \textbf{94.09} & \textbf{81.17} & \textbf{87.55} & \textbf{86.22} & \textbf{83.93}\\
\midrule
\color{gray}{Predictive Teacher }                                                                        &  $\checkmark$&-        & SwinTransformer                                                                    &   \color{gray}{88.07} & \color{gray}{99.45} & \color{gray}{93.79} & \color{gray}{90.41} & \color{gray}{83.60}& -        & -   \\
CFL+& $\times$&$\checkmark$                                                                                                                & SwinTransformer & {89.32} & {99.35} & \textbf{94.93} & {87.28} & 86.96 & {91.56} & {90.06}\\
\rowcolor{green!10} \ours             & $\times$&$\checkmark$  
                                                     & SwinTransformer & \textbf{89.83} & \textbf{99.41} & {94.85} & \textbf{88.10} & \textbf{87.26} & \textbf{91.89} & \textbf{90.53}\\
                                                     \midrule
 \color{gray}{Predictive Teacher }                                                                        &  $\checkmark$&-        & ConvNext-base                                                                  &  \color{gray}{89.14} & \color{gray}{99.46} & \color{gray}{94.36} & \color{gray}{91.77} & \color{gray}{87.37} & -        & -   \\
CFL+ &$\times$& $\checkmark$                                                                                                                & ConvNext-base & {87.95} & {98.98} & {91.71} & \textbf{87.65} & {72.45} & {87.75} & {87.02}\\
\rowcolor{green!10} \ours  & $\times$&$\checkmark$                                                         &ConvNext-base & {\textbf{88.11}} & \textbf{99.07} & \textbf{94.60} & \textbf{87.65} & \textbf{89.17} & \textbf{91.72} & \textbf{90.17}\\
\bottomrule
\end{tabular}%
}
\vspace{-1em}
\caption{
Impact of different architectures of the student. 
Predictive teachers are trained and tested on each dataset separately. 
}\label{tab: ablation: different student architecture}
\vspace{-1em}
\end{table*}

\noindent
\textbf{Robustness of \ours}.
To study the robustness of \ours
when the predictive teachers 
might be distracting to each other
due to the relevance between classes, we conduct experiments on the combination of 5 datasets, \ie, CUB200, Flower102, OxfordPet, Stanford Cars, and Dogs. 
As shown in \cref{tab: main 5 datasets},
our proposed \ours significantly outperforms existing methods, 
achieving the highest accuracy across all datasets except Stanford Dogs. 
This improvement is evident in both the average dataset accuracy and the average class accuracy, where \ours achieves 88.75\% and 86.15\%, respectively. 
However, we observe a performance drop on CUB200 and Dogs compared to the results in \cref{tab: ablation: 2 datasets}, which might be due to the noisy supervision.
We also notice that for Flowers102, Pets, Cars, and Dogs, \ours outperform the predictive teacher trained on each dataset with labels. 

\noindent
\textbf{Evaluation in a more challenging scenario}. 
To demonstrate the generalization capability of \ours,
we extended our experiments to include a total of 11 diverse datasets. 
Results in \Cref{tab: ablation: 10 datasets} show that \ours still outperforms baselines with a large margin on the average performance and most of the datasets. 
Meanwhile, we observe performance drops on 5 datasets compared to results in \Cref{tab: main 5 datasets}. 
This verifies the challenge of knowledge transfer from
teacher models trained on diverse data distribution and label spaces.
We also conclude that \ours better alleviates the performance drop 
caused by teachers' interference.

\subsection{Ablation Studies}
In this part, we present a series of ablation experiments using 5 datasets with the student model being ViT to demonstrate the effectiveness of different components of \ours. 
In this setting, we have $5\times 4=20$ predictive teachers and 60 descriptive teachers in total.

\begin{table*}[t!]
\centering
\resizebox{\textwidth}{!}{%
\begin{tabular}{l|c|ccccc|cc}
\toprule
\multirow{2}{*}{Methods} & \multirow{2}{*}{Loss for Feature Transfer} & \multicolumn{7}{c}{Accuracy}                                             \\
                         &                                                              & CUB200 & Flowers102 & Pets & Cars & Dogs & Avg. (D) & Avg. (C) \\
                         \midrule
                         
\multirow{2}{*}{\ours}&       smooth L1 + cosine distance (\cite{ranzinger2024radio}) &       78.20&96.66&\textbf{95.57}&40.95 & 68.13  & 75.90 & 70.81 \\
&        \cref{eqn: feature loss}  & \textbf{86.43} & \textbf{98.11} & {93.68} & \textbf{77.10} & \textbf{88.40} & \textbf{88.75} & \textbf{86.15}                         \\
\bottomrule
\end{tabular}%
}
\vspace{-1em}
\caption{
Ablation studies using the smooth L1 combined with the cosine distance (the feature loss from \cite{ranzinger2024radio}).
Avg. (D) and Avg. (C) refer to the average accuracy of datasets and classes. 
Best results are marked in \textbf{Bold}.}
\label{tab: ablation: loss}
\vspace{-1.5em}
\end{table*}

\paragraph{Impact of proposed voting-based modules.} 
To alleviate the conflicts between the features and logits of different teachers, we propose two novel voting mechanisms. 
Then, to understand how these methods contribute to the performance, we conduct experiments as shown in \Cref{tab: ablation: components}. 
The results demonstrate that integrating both components provides a significant boost in performance, achieving the highest accuracy on all datasets. 
Notably, the combined approach achieves 88.75\% and 86.15\% on the average dataset and class accuracy, surpassing the baseline CFL+. 
This underscores the importance of mitigating the conflicts among teachers for better knowledge summarization and transfer.

\paragraph{Impact of different types of public teachers. }
To understand how different types of public teachers contribute to the performance, we present an ablation study as shown in \cref{tab: ablation: different teachers}. 
The use of predictive teachers alone achieves a strong baseline, particularly on datasets like Flowers102 and Pets.
The combination of both types of teacher yields the highest average accuracies for both datasets and class levels. 
Moreover, compared with the baselines, \ie, CFL and CFL+, which employ predictive (and descriptive) teachers, \ours has significant performance gains. 
With predictive teachers, \ours improves performance from 83.17 to 87.88 on Avg. (D), while \ours boosts Avg. (D) from 87.21 to 88.75 with predictive and descriptive teachers.

\begin{figure}[t!]
\centering
    \begin{subfigure}[b]{\columnwidth}
        \centering
        \includegraphics[width=\columnwidth]{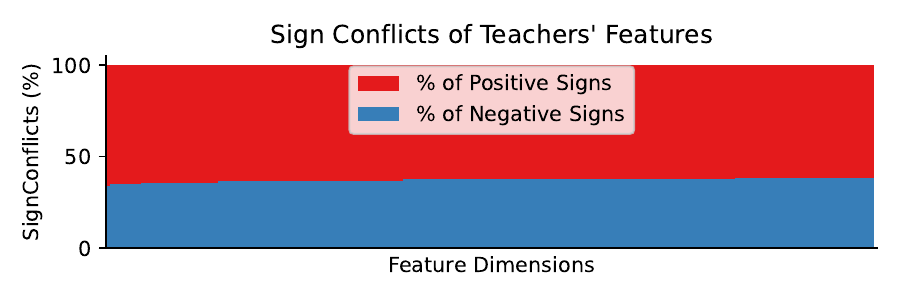}
        \vspace{-2em}
        \caption{An example of feature conflicts.}
    \end{subfigure}
    \begin{subfigure}[b]{\columnwidth}
        \centering
        \vspace{-0.2em}
\includegraphics[width=\columnwidth]{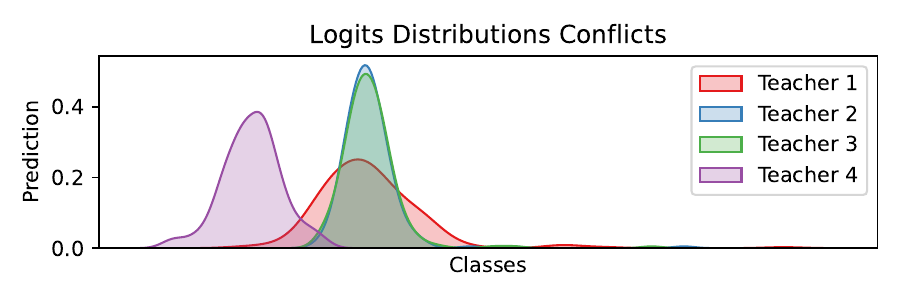}
        \vspace{-2em}
        \caption{An example of logit conflicts.}
    \end{subfigure}
    \vspace{-2.1em}
    \caption{A case study of feature and label conflicts.}
    \label{fig: case: conflicts}
    \vspace{-1.2em}
\end{figure}

\noindent
\textbf{Impact of the choice of student model.}
Our study also explores the impact of different student architectures including ResNet50~\cite{heDeepResidualLearning2016}, ViT-b32~\cite{dosovitskiyImageWorth16x162021}, ConvNeXt-base~\cite{liu_convnet_2022}, and SwinTransformer~\cite{liu_swin_2021}. 
The results are summarized in \Cref{tab: ablation: different student architecture}. 
Notably, Swin Transformer and ConvNext-base consistently demonstrate superior performance across all datasets, with Swin Transformer achieving the highest average dataset accuracy of 91.89\% and average class accuracy of 90.53\% under \ours. 
Similar results can be observed in the baseline (CFL+). 
It suggests that more advanced architectures can effectively capitalize on the diverse knowledge distilled from multiple teachers. 

\noindent
\textbf{Impact of different loss functions for feature transfer.}
To study the impact of different loss functions for feature transfer, 
we compare the proposed feature loss (\cref{eqn: feature loss}) 
with the combination of smooth L1 and cosine distance
which is commonly adopted by the state-of-the-art KD approaches 
like AM-Radio~\cite{ranzinger2024radio}.
The results in \Cref{tab: ablation: loss} show that our feature loss outperforms the combined loss~\cite{ranzinger2024radio} with a large margin on most of the datasets. 
However, we observe that the combined loss performs better on the Pets dataset, indicating that the combined loss might lead to an overfitting issue.

\noindent
\textbf{Qualitative analysis on feature and logit confilcts.}
To understand the feature and logit conflicts, we present a case study in \cref{fig: case: conflicts} using Stanford Cars. 
It shows that for feature conflicts, most of the teachers have negative signs, while for logit conflicts, the teachers have different opinions on the prediction. 
By simply averaging the features and logits, the conflicts will affect the student's performance as the features and logits are not consistent.

\subsection{How Well Does \ours Scale?}
To study how well does \ours scale, 
we present an analysis of increasing the number of datasets and teachers.
For the study about teachers number,
we present here only the results of ViT students under the 5-datasets setting
given the space limit.
The full results are in the Appendix.

\paragraph{Scaling on increasing the number of datasets.} 
We have presented the experimental results on 2 to 11 datasets in \Cref{tab: main 5 datasets,tab: ablation: 10 datasets,tab: ablation: 2 datasets}. 
Notably, \ours consistently outperforms all the baselines, which empirically proves that \ours can deal with hard scenarios where the label space overlap exists with highly diverse training data. 

\begin{figure}[t!]
    \begin{center}
        \subfloat[Avg. (D).\label{fig: teachers numbers: Dataset}]{\includegraphics[width=0.48\linewidth]{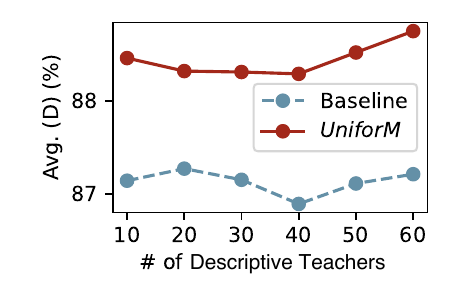}}
        \hfill
        \subfloat[Avg. (C).\label{fig: teachers numbers: Class}]{\includegraphics[width=0.48\linewidth]{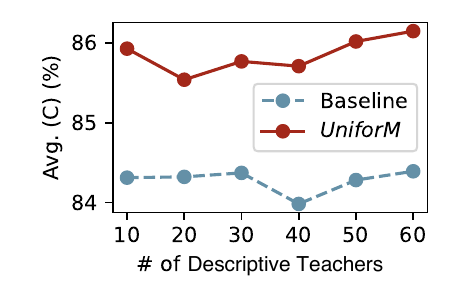}}

      \vspace{-1em}
        \caption{Performance when scaling up the number of public descriptive teachers under the 5-datasets setting. 
        More results are provided in the Appendix.
      }
        \label{tab: ablation: num of teachers}
   \end{center}
   \vspace{-2em}
\end{figure}

\paragraph{Scaling on increasing the number of public teachers.}
Here we consider controlling the number of descriptive teachers as most of the public teachers fall in this category. 
The results are presented in \Cref{tab: ablation: num of teachers}. 
It is obvious that increasing the number of descriptive teachers benefits as the performance improves. 
This shows a clear trend where introducing more teachers not only enriches the knowledge pool but also substantially boosts the student's performance with \ours.
However, we also notice that for baseline (CFL+), 
performance improvement saturates when increasing the number of teachers to 30. 
We hypothesize this is because the sign and logit distribution conflicts bring noise to the logit and feature transfer, limiting the improvement, which also shows the superiority of \ours in handling noisy knowledge.

\section{Conclusion}

In this paper, we introduced \textit{\ours}, a pioneering framework for unifying knowledge transfer from diverse online public pre-trained models with heterogeneous architectures and training data. 
\ours had two novel voting-based components, \ie, voting for feature transfer and voting for logit transfer, which effectively handled the heterogeneity and conflicts in feature and logit distributions that typically hinder conventional knowledge distillation approaches. 
Through extensive experimental validation across 11 benchmark datasets, with 104 public teacher models, \textit{\ours} significantly outperformed existing baselines.

\noindent
\textbf{Limitations.} 
It will be worth conducting experiments with more benchmark datasets and public teachers to evaluate the effectiveness of \ours further. 
Moreover, with the increase of teachers and benchmark datasets, it will be interesting to deal with the huge discrepancy between them and leverage it for better knowledge transfer.
Last, we only evaluate on image classification. 
It is worth exploring efficient knowledge integration methods on other vision tasks.

{
    \small
    \bibliographystyle{ieeenat_fullname}
    \bibliography{references,new}
}

\clearpage
\setcounter{page}{1}
\maketitlesupplementary

\input{sections/6_appendix}

\end{document}

%% file: utils.tex
\usepackage[utf8]{inputenc} %
\usepackage[T1]{fontenc}    %
\usepackage{hyperref}       %
\usepackage{url}            %
\usepackage{booktabs}       %
\usepackage{amsfonts}       %
\usepackage{nicefrac}       %
\usepackage{microtype}      %
\usepackage[dvipsnames]{xcolor}         %
\usepackage{soul}

\usepackage{subcaption}

\usepackage{amsmath}
\usepackage{graphicx}

\usepackage{colortbl}

\usepackage{booktabs}
\usepackage{multirow}

\usepackage{soul}

\newcommand{\ours}{\textsc{Uniform}\xspace}

%% file: sections/6_appendix.tex
In this technical Appendix, we present details on the experiments, including baselines and public teachers.

\section{Experiments}

\subsection{How Well Does \ours Scale - Number of Descriptive Teachers}

We present the visualization of the performance of \ours when scaling up the number of public descriptive teachers in \Cref{tab: ablation: num of teachers full}. 
The results are consistent with the findings in the main paper, demonstrating that \ours can effectively leverage the knowledge from a large number of descriptive teachers to improve the student's performance. 
Though we found that the performance tends to saturate when the number of teachers exceeds 30 on some datasets, the performance continues to improve with more teachers on other datasets.

\subsection{Details of Baselines}

To compare the effectiveness of \ours, we establish two types of baselines, detailed below:

\begin{figure}[t!]
    \begin{center}
        \subfloat[Avg. (D).\label{fig: teachers numbers: Dataset}]{\includegraphics[width=0.48\linewidth]{figures/Avg.D.pdf}}
        \hfill
        \subfloat[Avg. (C).\label{fig: teachers numbers: Class}]{\includegraphics[width=0.48\linewidth]{figures/Avg.C.pdf}}
        
        \subfloat[CUB200.\label{fig: teachers numbers: CUB200}]{\includegraphics[width=0.48\linewidth]{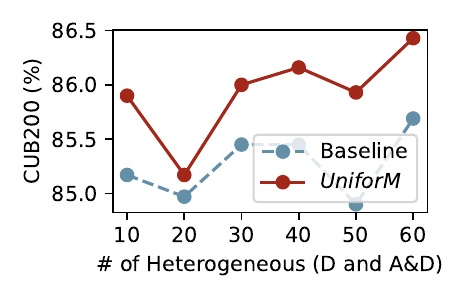}}
        \hfill
        \subfloat[Flowers102.\label{fig: teachers numbers: Flowers}]{\includegraphics[width=0.48\linewidth]{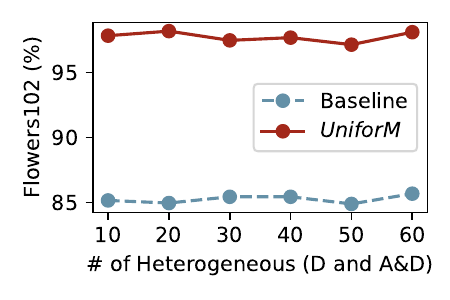}}

        \subfloat[OxfordPet.\label{fig: teachers numbers: Pets}]{\includegraphics[width=0.48\linewidth]{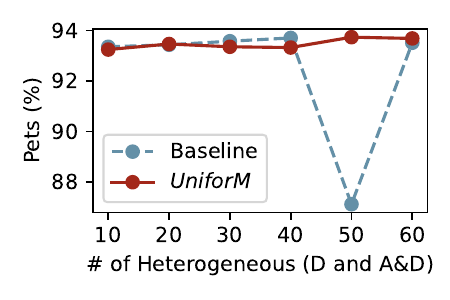}}
        \hfill
        \subfloat[Stanford Cars.\label{fig: teachers numbers: Cars}]{\includegraphics[width=0.48\linewidth]{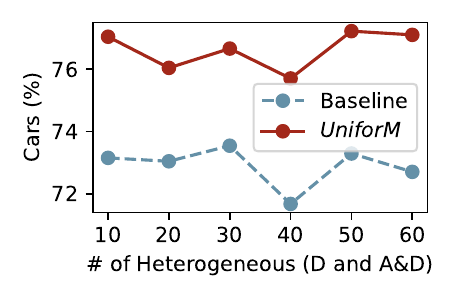}}

        \subfloat[Stanford Dogs.\label{fig: teachers numbers: Dogs}]{\includegraphics[width=0.48\linewidth]{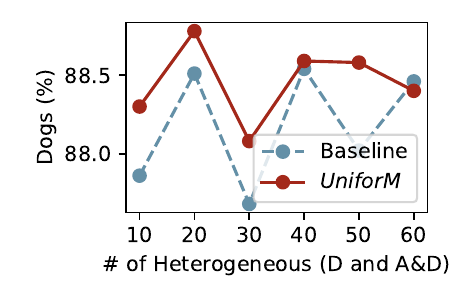}}
        
        \caption{Performance when scaling up the number of public descriptive teachers under the 5 datasets setting (main configuration).
      }
        \label{tab: ablation: num of teachers full}
   \end{center}
\end{figure}

\noindent
\textbf{Methods with predictive teachers only}. 
We introduce three baselines, \ie, Knowledge Distillation (KD)~\cite{hinton_distilling_2015}, CFL~\cite{luo_knowledge_2019}, and OFA~\cite{hao_one-for-all_2023} without ground-truth labels. 
These methods learn solely from predictive teachers without relying on ground-truth labels.

\textbf{Knowledge Distillation (KD)}. 
In this baseline, we employ logit and feature transfer from homogeneous teachers, which are trained on the same dataset and share the same architecture as the student. 
Specifically, we denote the logits and features from $N^{homo}$ teachers as $\{\bm{x}^{t}_{i,j}\}_{j\in[N^{homo}]}$ and $\{\bm{p}^{t}_{i,j}\}_{j\in[N^{homo}]}$ for the $i$-th input, respectively.  
The average feature $\bar{\bm{x}}^{t}_{i}$ and logit $\bar{\bm{p}}^{t}_{i}$ serve as the training targets for the $i$-th input, 
\begin{equation}
    \begin{aligned}
        \bar{\bm{x}}^{t}_{i} =& \frac{\sum_{j\in[N^{homo}]} \bm{x}^{t}_{i,j}}{N^{homo}}\,,\\
        \bar{\bm{p}}^{t}_{i} =& \frac{\sum_{j\in[N^{homo}]} \bm{p}^{t}_{i,j}}{N^{homo}}\,.
    \end{aligned}
\end{equation}
The loss used in KD is
\begin{equation}
    \ell_{KD} = dist(\bm{x}, \bar{\bm{x}}^{t}) + \sum_{c\in[C]} \bar{p}^t_{c} \log p_{c}\,,
\end{equation}
where $p_c$ is the $c$-th element of $\bm{p}$.

\textbf{CFL.} 
We adapt CFL to handle multiple teacher scenarios, incorporating all predictive teachers (both homogeneous and heterogeneous (A) types), learning from the average features and logits as KD. 
The hyperparameters remain consistent with those in the original paper.

\textbf{OFA.}
For OFA, we modify its design by removing the cross-entropy loss from Equation (1) of the original paper while keeping all other details unchanged. 
OFA also learns from the average features and logits.

\noindent
\textbf{Methods with predictive and descriptive teachers}: Since there does not exist a method that can learn from predictive and descriptive teachers simultaneously, we extend CFL to support the descriptive teachers as our baseline, referred to as CFL+. 

CFL+ is specifically designed to accommodate descriptive teachers, \ie, heterogeneous (D and A\&D) teachers, and adopts the same encoder and decoder structure ($f^e(\cdot)$ and $f^d(\cdot)$) as \ours. 
 and $f^d(\cdot)$) as \ours. 
However, different from \ours, CFL+ directly averages the features and logits from all public teachers to serve as the student's targets, similar to KD. 
Specifically, CFL+ employs the following loss,
\begin{equation}
    \begin{aligned}
        \ell_{logit} = & \sum_{c\in[C]}\bar{p}^{t}_c \log p_c \,, \\
        \ell_{feature} = & dist(\bar{\bm{x}}^{t}, \bar{\bm{x}})\,, \\
        \ell_{CFL+} = & \ell_{logit} + \beta_{1} \ell_{feature} + \beta_{2}\ell_{rec}\,,
        \,
    \end{aligned}
\end{equation}
where $\beta_1$ and $\beta_2$ are the same with \ours.

In all baseline methods, the label spaces of the logits are unified across different datasets. 
When combining multiple datasets, the label spaces of all datasets are merged into a single, unified space. 
This ensures that logits from all predictive teachers align with the same set of classes, enabling fair comparisons with \ours, which also operates in a unified label space. 
Specifically, if there are $K$ datasets, each with a label space 
$\mathcal{Y}_i = \{y_c\}_{c \in [C_i]}$, the \textbf{unified label space} $\mathcal{Y}$ is the union of all individual label spaces $\mathcal{Y} = \bigcup_{i=1}^{K} \mathcal{Y}_i$.
Each sample is assigned a target in the combined label space, and logits from all predictive teachers are aligned with this unified set of classes. For a student learning from $K$ datasets, the logits $\bm{p}_i^t$ from the $i$-th teacher are projected (reassigning indices and padding with zero) to this unified space, ensuring consistency across datasets.

\subsection{Details of Public Teachers}

\begin{table}[t!]
\centering
\resizebox{\columnwidth}{!}{%
\begin{tabular}{l|cccc}
\toprule
 & ResNet50 & ViT-b32 & ConvNeXt-base & SwinTransformer \\
\midrule
CUB200 & \href{https://huggingface.co/anonauthors/cub200-resnet50}{link} & \href{https://huggingface.co/anonauthors/cub200-ViT-b32}{link} &\href{https://huggingface.co/anonauthors/cub200-ConvNeXt-base}{link} & \href{https://huggingface.co/anonauthors/cub200-swinT}{link} \\
Flower102 & \href{https://huggingface.co/anonauthors/flowers102-resnet50}{link} & \href{https://huggingface.co/anonauthors/flowers102-ViT-b32}{link} & \href{https://huggingface.co/anonauthors/flowers102-ConvNeXt-base}{link} & \href{https://huggingface.co/anonauthors/flowers102-swinT}{link} \\
Oxford Pets & \href{https://huggingface.co/anonauthors/oxford\_pet-resnet50}{link} & \href{https://huggingface.co/anonauthors/oxford\_pet-ViT-b32}{link} & \href{https://huggingface.co/anonauthors/oxford\_pet-ConvNeXt-base}{link}  & \href{https://huggingface.co/anonauthors/oxford\_pet-swinT}{link} \\
Stanford Cars & \href{https://huggingface.co/anonauthors/stanford\_cars-resnet50}{link} & \href{https://huggingface.co/anonauthors/stanford\_cars-ViT-b32}{link} & \href{https://huggingface.co/anonauthors/stanford\_cars-ConvNeXt-base}{link} & \href{https://huggingface.co/anonauthors/stanford\_cars-swinT}{link} \\
Stanford Dogs & \href{https://huggingface.co/anonauthors/stanford\_dogs-resnet50}{link}   & \href{https://huggingface.co/anonauthors/stanford\_dogs-ViT-b32}{link} & \href{https://huggingface.co/anonauthors/stanford\_dogs-ConvNeXt-base}{link} & \href{https://huggingface.co/anonauthors/stanford\_dogs-swinT}{link}  \\
Food101  & \href{https://huggingface.co/anonauthors/food101-resnet50}{link}  & \href{https://huggingface.co/anonauthors/food101-ViT-b32}{link}  & \href{https://huggingface.co/anonauthors/food101-ConvNeXt-base}{link} & \href{https://huggingface.co/anonauthors/food101-swinT}{link} \\
Caltech101  & \href{https://huggingface.co/anonauthors/caltech101-resnet50}{link}  & \href{https://huggingface.co/anonauthors/caltech101-ViT-b32}{link}     & \href{https://huggingface.co/anonauthors/caltech101-ConvNeXt-base}{link} & \href{https://huggingface.co/anonauthors/caltech101-swinT}{link}      \\
Cifar10  & \href{https://huggingface.co/anonauthors/cifar10-resnet50}{link}    & \href{https://huggingface.co/anonauthors/cifar10-ViT-b32}{link}    & \href{https://huggingface.co/anonauthors/cifar10-ConvNeXt-base}{link}     & \href{https://huggingface.co/anonauthors/cifar10-swinT}{link}            \\
Cifar100   & \href{https://huggingface.co/anonauthors/cifar100-resnet50}{link}     & \href{https://huggingface.co/anonauthors/cifar100-ViT-b32}{link}      & \href{https://huggingface.co/anonauthors/cifar100-ConvNeXt-base}{link}      & \href{https://huggingface.co/anonauthors/cifar100-swinT}{link}           \\
DTD      & \href{https://huggingface.co/anonauthors/dtd-resnet50}{link}           & \href{https://huggingface.co/anonauthors/dtd-ViT-b32}{link}     & \href{https://huggingface.co/anonauthors/dtd-ConvNeXt-base}{link}                & \href{https://huggingface.co/anonauthors/dtd-swinT}{link}             \\
Aircraft     & \href{https://huggingface.co/anonauthors/fgvc\_aircraft-resnet50}{link}     & \href{https://huggingface.co/anonauthors/fgvc\_aircraft-ViT-b32}{link}     & \href{https://huggingface.co/anonauthors/fgvc\_aircraft-ConvNeXt-base}{link}         & \href{https://huggingface.co/anonauthors/fgvc\_aircraft-swinT}{link}       \\
\bottomrule
\end{tabular}%
}
\caption{The links to public predictive teachers.}
\label{tab:homogeneous teachers}
\end{table}

The predictive and descriptive teachers used in our experiments are detailed in \Cref{tab: selected public models,tab: HF public models,tab:homogeneous teachers}. 

For the \textbf{predictive teachers}, we select four base models per dataset, resulting in a total of 44 predictive teachers across the 11 datasets, as listed in \Cref{tab:homogeneous teachers}. 

For the \textbf{descriptive teachers}, we aim to leverage both well-known representative models and widely used public models. Specifically:
\begin{itemize}
    \item We include 10 representative models chosen based on expert knowledge, as detailed in \Cref{tab: selected public models}.
    \item Additionally, we select the top 50 most downloaded models from HuggingFace, as shown in \Cref{tab: HF public models}.
\end{itemize}

In total, for the experiments conducted on 11 datasets, we utilize $44 + 10 + 50 = 104$ teachers.

\begin{table}[t]
\centering
\begin{tabular}{ll}
\toprule
\multicolumn{1}{c}{\#} & \multicolumn{1}{c}{Model Name} \\
1  & VGG19 \\
2  & ResNet18 \\
3  & ResNet34 \\
4  & ResNet50 \\
5  & ResNet101 \\
6  & ConvNext-Base \\
7  & ViT-B32 \\
8  & DeIT-S16 \\
9  & ResMLP-12 \\
10  & SwinTransfromer-Base \\
\bottomrule
\end{tabular}
\caption{Selected representative public descriptive teachers.}
\label{tab: selected public models}
\end{table}
\begin{table}[t!]
\centering
\resizebox{\columnwidth}{!}{%
\begin{tabular}{lll}
\toprule
Type                                        & \multicolumn{1}{c}{\#} & \multicolumn{1}{c}{Model Name and Link} \\
\midrule
\multirow{50}{*}{\rotatebox{90}{HuggingFace Top Downloaded}} & 1 & \href{https://huggingface.co/timm/resnet50.a1_in1k}{timm/resnet50.a1-in1k} \\
 & 2    &  \href{https://huggingface.co/google/vit-base-patch16-224}{google/vit-base-patch16-224}  \\
 & 3    &  \href{https://huggingface.co/timm/mobilenetv3_large_100.ra_in1k}{timm/mobilenetv3-large-100.ra-in1k}  \\
 & 4    &  \href{https://huggingface.co/microsoft/beit-base-patch16-224-pt22k-ft22k}{microsoft/beit-base-patch16-224-pt22k-ft22k}  \\
 & 5    &  \href{https://huggingface.co/timm/resnet18.a1_in1k}{timm/resnet18.a1-in1k}  \\
 & 6    &  \href{https://huggingface.co/amunchet/rorshark-vit-base?library=true}{amunchet/rorshark-vit-base}  \\
 & 7    &  \href{https://huggingface.co/rizvandwiki/gender-classification?library=true}{rizvandwiki/gender-classification}  \\
 & 8    & \href{https://huggingface.co/timm/convnext_small.fb_in22k}{timm/convnext-small.fb-in22k}   \\
 & 9    &  \href{https://huggingface.co/nvidia/mit-b0}{nvidia/mit-b0}  \\
 & 10    &  \href{https://huggingface.co/timm/resnet18.fb_swsl_ig1b_ft_in1k}{timm/resnet18.fb-swsl-ig1b-ft-in1k}  \\
 &11 &\href{https://huggingface.co/trpakov/vit-face-expression}{trpakov/vit-face-expression}\\
&12 &\href{https://huggingface.co/timm/nfnet_l0.ra2_in1k}{timm/nfnet-l0.ra2-in1k}\\
&13 & \href{https://huggingface.co/timm/vit_tiny_patch16_224.augreg_in21k_ft_in1k}{timm/vit-tiny-patch16-224.augreg-in21k-ft-in1k} \\
&14 & \href{https://huggingface.co/timm/swin_base_patch4_window7_224.ms_in22k_ft_in1k}{timm/swin-base-patch4-window7-224.ms-in22k-ft-in1k} \\
&15 & \href{https://huggingface.co/nateraw/vit-age-classifier}{nateraw/vit-age-classifier} \\
&16 & \href{https://huggingface.co/facebook/convnext-tiny-224}{facebook/convnext-tiny-224} \\
&17 & \href{https://huggingface.co/NTQAI/pedestrian_gender_recognition?library=true}{NTQAI/pedestrian-gender-recognition} \\
&18 & \href{https://huggingface.co/Kaludi/food-category-classification-v2.0?library=true}{Kaludi/food-category-classification-v2.0} \\
&19 &\href{https://huggingface.co/nateraw/food?library=true}{nateraw/food} \\
&20 & \href{https://huggingface.co/timm/vgg19.tv_in1k}{timm/vgg19.tv-in1k}\\
&21 &\href{https://huggingface.co/timm/vit_base_patch16_224.augreg_in21k}{timm/vit-base-patch16-224.augreg-in21k}\\
&22 &\href{https://huggingface.co/timm/tf_efficientnet_b0.ns_jft_in1k}{timm/tf-efficientnet-b0.ns-jft-in1k} \\
&23 & \href{https://huggingface.co/timm/mobilenetv2_100.ra_in1k}{timm/mobilenetv2-100.ra-in1k} \\
&24 & \href{https://huggingface.co/microsoft/dit-base-finetuned-rvlcdip}{microsoft/dit-base-finetuned-rvlcdip} \\
&25 & \href{https://huggingface.co/apple/mobilevit-small}{apple/mobilevit-small} \\
&26 & \href{https://huggingface.co/timm/resnetv2_50x1_bit.goog_in21k}{timm/resnetv2-50x1-bit.goog-in21k} \\
&27 & \href{https://huggingface.co/aaraki/vit-base-patch16-224-in21k-finetuned-cifar10?library=true}{aaraki/vit-base-patch16-224-in21k-finetuned-cifar10}\\
&28 & \href{https://huggingface.co/Falconsai/nsfw_image_detection}{Falconsai/nsfw-image-detection} \\
&29 & \href{https://huggingface.co/timm/resnet101.a1h_in1k}{timm/resnet101.a1h-in1k} \\
&30 & \href{https://huggingface.co/timm/efficientnet_b0.ra_in1k}{timm/efficientnet-b0.ra-in1k} \\
&31 & \href{https://huggingface.co/google/mobilenet_v2_1.0_224}{google/mobilenet-v2-1.0-224} \\
&32 & \href{https://huggingface.co/WinKawaks/vit-tiny-patch16-224?library=true}{WinKawaks/vit-tiny-patch16-224} \\
&33 & \href{https://huggingface.co/timm/fbnetc_100.rmsp_in1k}{timm/fbnetc-100.rmsp-in1k} \\
&34 & \href{https://huggingface.co/timm/tf_efficientnetv2_s.in21k}{timm/tf-efficientnetv2-s.in21k} \\
&35 & \href{https://huggingface.co/google/mobilenet_v1_0.75_192}{google/mobilenet-v1-0.75-192} \\
&36 & \href{https://huggingface.co/timm/convnext_base.fb_in22k_ft_in1k}{timm/convnext-base.fb-in22k-ft-in1k} \\
&37 & \href{https://huggingface.co/microsoft/beit-base-patch16-224}{microsoft/beit-base-patch16-224} \\
&38 & \href{https://huggingface.co/timm/vit_small_patch16_224.augreg_in21k_ft_in1k}{timm/vit-small-patch16-224.augreg-in21k-ft-in1k} \\
&39 & \href{https://huggingface.co/timm/hrnet_w18.ms_aug_in1k}{timm/hrnet-w18.ms-aug-in1k} \\
&40 & \href{https://huggingface.co/timm/mobilevit_s.cvnets_in1k}{timm/mobilevit-s.cvnets-in1k} \\
&41 & \href{https://huggingface.co/timm/convnextv2_atto.fcmae_ft_in1k}{timm/convnextv2-atto.fcmae-ft-in1k}\\
&42 & \href{https://huggingface.co/timm/cspdarknet53.ra_in1k}{timm/cspdarknet53.ra-in1k} \\
&43 & \href{https://huggingface.co/microsoft/swin-tiny-patch4-window7-224}{microsoft/swin-tiny-patch4-window7-224} \\
&44 & \href{https://huggingface.co/timm/ese_vovnet19b_dw.ra_in1k}{timm/ese-vovnet19b-dw.ra-in1k} \\
&45 & \href{https://huggingface.co/timm/tf_efficientnetv2_s.in21k_ft_in1k}{timm/tf-efficientnetv2-s.in21k-ft-in1k} \\
&46 & \href{https://huggingface.co/timm/mixer_b16_224.goog_in21k_ft_in1k}{timm/mixer-b16-224.goog-in21k-ft-in1k}\\
&47 & \href{https://huggingface.co/timm/deit_base_distilled_patch16_224.fb_in1k}{timm/deit-base-distilled-patch16-224.fb-in1k} \\
&48 & \href{https://huggingface.co/timm/pnasnet5large.tf_in1k}{timm/pnasnet5large.tf-in1k} \\
&49 & \href{https://huggingface.co/timm/pit_b_224.in1k}{timm/pit-b-224.in1k}\\
&50 & \href{https://huggingface.co/timm/mnasnet_100.rmsp_in1k}{timm/mnasnet-100.rmsp-in1k} \\
\bottomrule
\end{tabular}
}
\caption{Top 50 downloaded vision models on HuggingFace we use for public models. This might be different from HuggingFace as time changes.}
\label{tab: HF public models}

\end{table}